\documentclass[10pt,twocolumn,letterpaper]{article}

\usepackage{subfigure}
\usepackage{cvpr}
\usepackage{times}
\usepackage{epsfig}
\usepackage{graphicx}
\usepackage{amsmath}
\usepackage{amssymb}
\usepackage{booktabs}
\usepackage{multirow}
\usepackage{url}
\usepackage{cite}
\newcommand{\Paragraph}[1]{\vspace{-0mm} \noindent \textbf{#1.} \hspace{0mm}}

\usepackage{subfiles}

\usepackage[pagebackref=true,breaklinks=true,letterpaper=true,colorlinks,bookmarks=false]{hyperref}

 \cvprfinalcopy % *** Uncomment this line for the final submission

\def\cvprPaperID{4755} % *** Enter the CVPR Paper ID here

\def\figvspace{\vspace{-4mm}}

\ifcvprfinal\fi
\begin{document}

%%%%%%%%% TITLE
\title{On the Detection of Digital Face Manipulation}

\author{Hao Dang\thanks{denotes equal contribution by the authors. } $\quad$ Feng Liu\footnotemark[1] $\quad$ Joel Stehouwer\footnotemark[1] $\quad$ Xiaoming Liu $\quad$ Anil Jain\\
Department of Computer Science and Engineering\\
Michigan State University, East Lansing MI 48824\\
}

\maketitle
\thispagestyle{empty}

%%%%%%%%% ABSTRACT
\begin{abstract}

Detecting manipulated facial images and videos is an increasingly important topic in digital media forensics.
As advanced face synthesis and manipulation methods are made available, new types of fake face representations are being created which have raised significant concerns for their use in social media.
Hence, it is crucial to detect manipulated face images and localize manipulated regions.
Instead of simply using multi-task learning to simultaneously detect manipulated images and predict the manipulated mask (regions), we propose to utilize an attention mechanism to process and improve the feature maps for the classification task.
The learned attention maps highlight the informative regions to further improve the binary classification (genuine face v. fake face), and also visualize the manipulated regions.
To enable our study of manipulated face detection and localization, we collect a large-scale database that contains numerous types of facial forgeries.
With this dataset, we perform a thorough analysis of data-driven fake face detection. 
We show that the use of an attention mechanism improves facial forgery detection and manipulated region localization.
The code and database are available at \url{cvlab.cse.msu.edu/project-ffd.html}.

\end{abstract}

%%%%%%%%% BODY TEXT
%--------------------------------------------------------------------%
\section{Introduction}
\label{sec:intro}
Human faces play an important role in human-human communication and association of side information, {\it e.g.}, gender and age with identity.
For instance, face recognition is increasingly utilized in our daily life for applications such as access control and payment~\cite{representation-learning-by-rotating-your-faces}.
However, these advances also entice malicious actors to manipulate face images to launch attacks, aiming to be authenticated as the genuine user.
Moreover, manipulation of facial content has become ubiquitous, and raises new concerns especially in social media content~\cite{shu2017fake, rossler2018faceforensics, rossler2019faceforensics++}.
Recent advances in deep learning have led to a dramatic increase in the realism of face synthesis and enabled a rapid dissemination of ``fake news''~\cite{agarwal2019protecting}.
Therefore, to mitigate the adverse impact and benefit both \textit{public security and privacy}, it is crucial to develop effective solutions against these facial forgery attacks.

\begin{figure}[t]
\begin{center}
\includegraphics[width=0.99\linewidth]{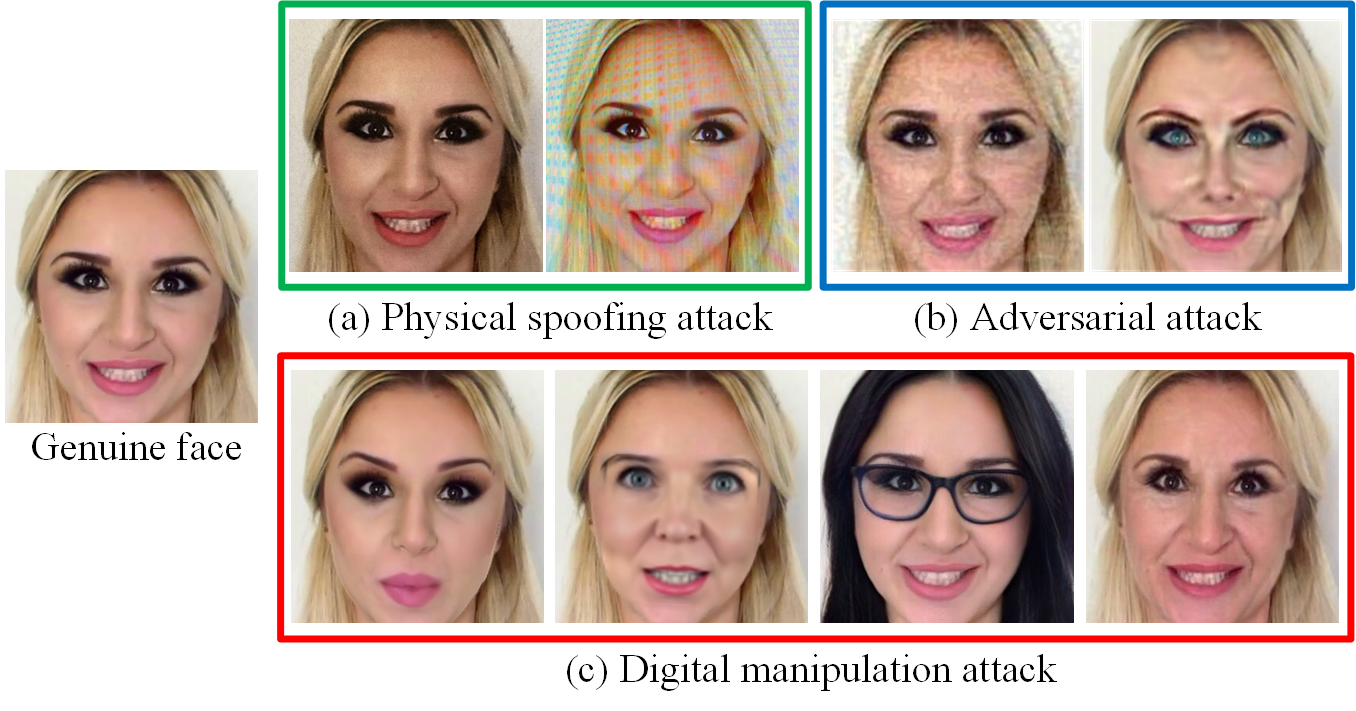}
\end{center}
\figvspace
%\vspace{-1mm}
\caption{\small Given one genuine face image, there are three types of facial forgery attacks: physical spoofing attack (print and replay attack), adversarial attack~\cite{deb2019advfaces}, and digital manipulation attack.}
\label{fig:three_types}
\figvspace
\end{figure}

As shown in Fig.~\ref{fig:three_types}, there are \emph{three} main types of facial forgery attacks.
i) Physical spoofing attacks can be as simple as face printed on a paper, replaying image/video on a phone, or as complicated as a $3$D mask~\cite{face-anti-spoofing-using-patch-and-depth-based-cnns,liu2018learning, jourabloo2018face, liu2019deep}.
ii) Adversarial face attacks generate high-quality and perceptually imperceptible adversarial images that can evade automated face matchers~\cite{goodfellow2014explaining, madry2017towards, deb2019advfaces,xu2019adversarial}.
iii) Digital manipulation attacks, made feasible by Variational AutoEncoders (VAEs)~\cite{kingma2013auto, rezende2014stochastic} and Generative Adversarial Networks (GANs)~\cite{goodfellow2014generative}, can generate entirely or partially modified photorealistic face images.
Among these three types, this work addresses only \emph{digital manipulation attacks}, with the objectives of automatically detecting manipulated faces, as well as localizing modified facial regions.
We  use the term ``face manipulation detection" or ``face forgery detection" to describe our objective.

\begin{figure}[t]
\begin{center}
\includegraphics[width=\linewidth]{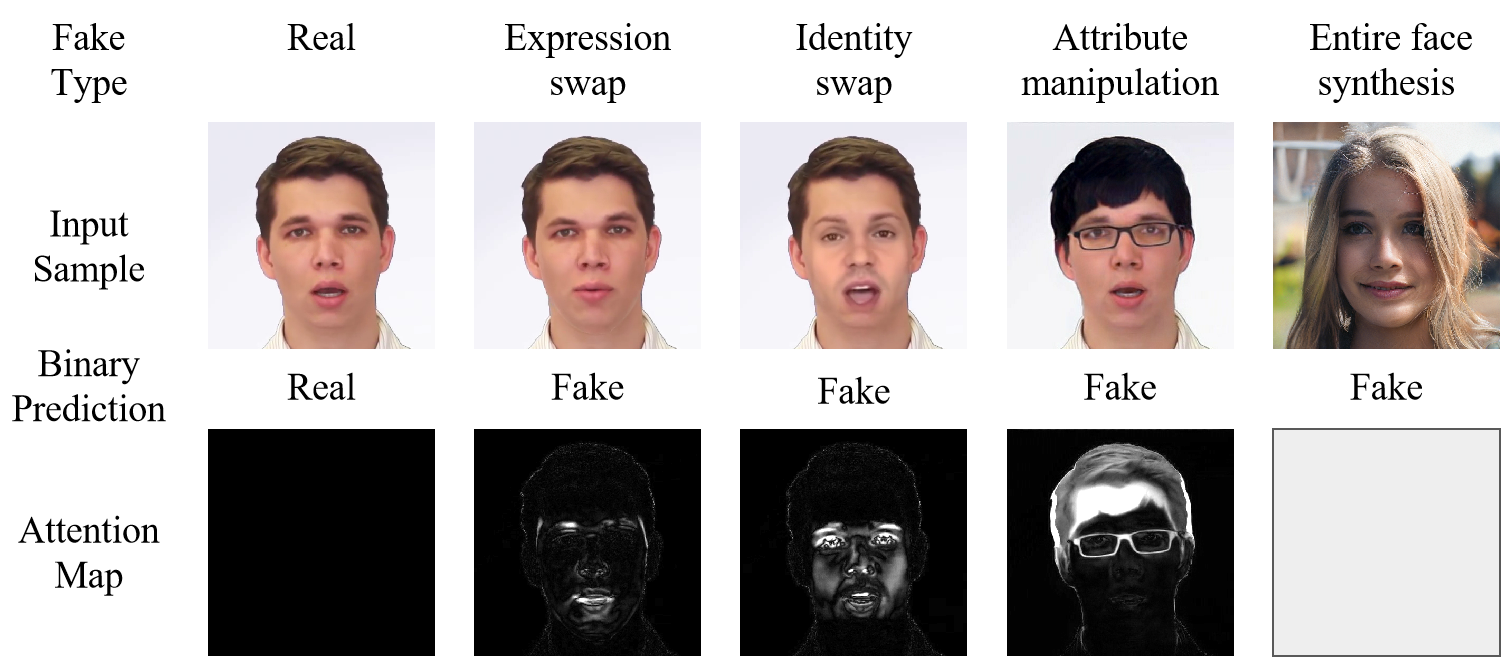}
\figvspace
\vspace{-2mm}
\end{center}
\caption{\small Our facial forgery detection method tackles faces generated by the four types of face manipulation methods. Given a face image, our approach outputs a binary decision (genuine v. manipulated face) and localizes the manipulated regions via an estimated attention map. For real or entirely synthetic faces, our estimated maps are assumed to be uniformly distributed in [0, 1].}
\label{fig:overview}
\figvspace
\end{figure}

Digital facial manipulation methods fall into four categories: expression swap, identity swap, attribute manipulation and entire face synthesis (Fig.~\ref{fig:overview}).
$3$D face reconstruction and animation methods~\cite{dale2011video, zollhofer2018state,Liu_2018_CVPR,on-learning-3d-face-morphable-model-from-in-the-wild-images} are widely used for expression swap, such as \emph{Face2Face}~\cite{thies2016face2face}.
These methods can transfer expressions from one person to another in real time with only RGB cameras.
Identity swap methods replace the face of one person with the face of another.
Examples include \emph{FaceSwap}~\cite{thies2016face2face,wang2019detecting}, which inserts famous actors into movie clips in which they never appeared and \emph{DeepFakes}~\cite{deepfakes}, which performs face swapping via deep learning algorithms.
 
Attribute manipulation edits single or multiple attributes in a face, \emph{e.g.}, gender, age, skin color, hair, and glasses.
The adversarial framework of GANs is used for image translation~\cite{isola2017image, zhu2017unpaired, zhu2017toward} or manipulation in a given context~\cite{bau2018gan, suwajanakorn2017synthesizing}, which diversifies facial images synthesis.
\emph{FaceApp}~\cite{FaceApp} has popularized facial attribute manipulation as a consumer-level application, providing $28$ filters to modify specific attributes~\cite{FaceApp}.
The fourth category is entire face synthesis.
Fueled by the large amounts of face data and the success of GANs, any user is capable of producing a completely synthetic facial image, whose realism is such that even humans have difficulty assessing if it is genuine or manipulated~\cite{karras2017progressive, choi2018stargan, karras2019style}.

Research on face manipulation detection has been seriously hampered by the lack of large-scale datasets of manipulated faces.
Existing approaches are often evaluated on small datasets with limited manipulation types, including Zhou~\etal~\cite{zhou2017two}, Deepfake~\cite{korshunov2018deepfakes}, and FaceForensics/FaceForensics${++}$~\cite{rossler2018faceforensics, rossler2019faceforensics++}.
To remedy this issue, we collect a Diverse Fake Face Dataset (DFFD) of $2.6$ million images from all four categories of digital face manipulations.

\begin{figure*}[t]
\begin{center}
\includegraphics[width=0.95\linewidth]{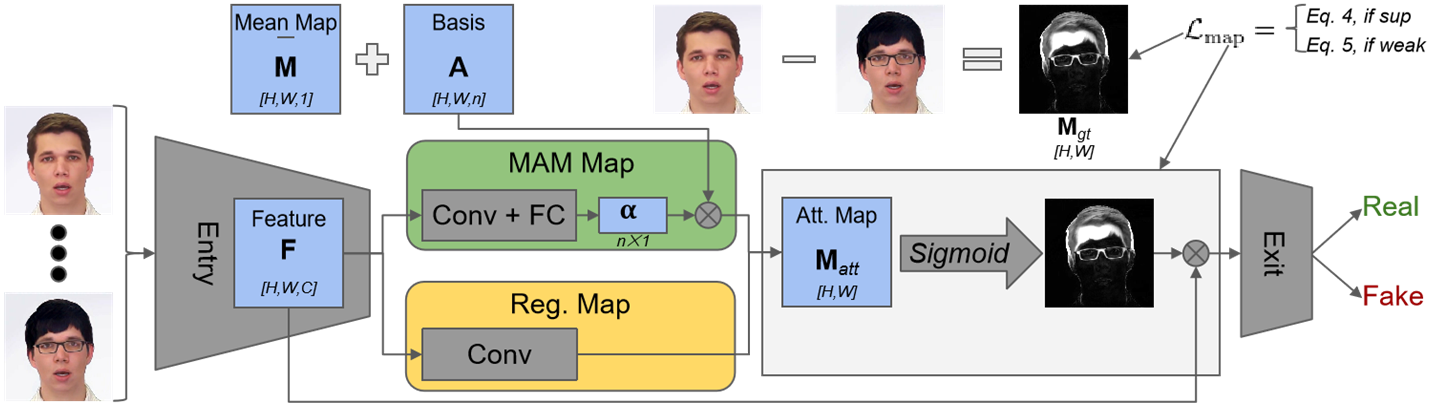}

\vspace{-3mm}
\end{center}
\caption{\small The architecture of our face manipulation detection.
Given any backbone network, our proposed attention-based layer can be inserted into the network.
It takes the high-dimensional feature $\mathbf{F}$ as input, estimates an attention map $\mathbf{M}_{att}$ using either {\it MAM}-based or {\it regression}-based methods, and channel-wise multiplies it with the high-dimensional features, which are fed back into the backbone. In addition to the binary classification supervision $\mathcal{L}_{\text{classifier}}$, either a supervised or weakly supervised loss, $\mathcal{L}_{\text{map}}$, can be applied to estimate the attention map, depending on whether the ground truth manipulation map $\mathbf{M}_{gt}$ is available.}
\label{fig:network_architecture}
\figvspace
%FIXME 
%The output of Sigmoid is NOT M_att. L_map is applied between M_gt and M_att (before the Sigmoid). 
\end{figure*}

Due to the fact that the modification of a face image can be in whole or in part, we assume that a well-learned network would gather different information {\it spatially} in order to detect manipulated faces.
We hypothesize that correctly estimating this spatial information can enable the network to focus on these important spatial regions to make its decision.
Hence, we aim to not only detect manipulated faces, but also automatically locate the manipulated regions by estimating an image-specific attention map, as in Fig.~\ref{fig:network_architecture}.
We present our approach to estimate the attention map in both supervised and weakly-supervised fashions.
We also demonstrate that this attention map is beneficial to the final task of facial forgery detection.
Finally, in order to quantify the attention map estimation, we propose a novel metric for attention map accuracy evaluation.
In the future, we anticipate the predicted attention maps for manipulated face images and videos would reveal hints about the type, magnitude, and even intention of the manipulation.

In summary, the contributions of this work include:

$\diamond$ A comprehensive fake face dataset including $0.8$M real and $1.8$M fake faces generated by a diverse set of face modification methods and an accompanying evaluation protocol.

$\diamond$ A novel attention-based layer to improve classification performance and produce an attention map indicating the manipulated facial regions. 

$\diamond$ A novel metric, termed Inverse Intersection Non-Containment (IINC), for evaluating attention maps that produces a more coherent evaluation than existing metrics.

$\diamond$ State-of-the-art performance of digital facial forgery detection for both seen and unseen manipulation methods.

\section{Related Work}

\Paragraph{Digital Face Manipulation Methods} With the rapid progress in computer graphics and computer vision,  it is becoming difficult for humans to tell the difference between genuine and manipulated faces~\cite{rossler2019faceforensics++}.
Graphics-based approaches are widely used for identity or expression transfer by first reconstructing $3$D models for both source and target faces, and then exploiting the corresponding $3$D geometry  to warp between them.
In particular, Thies~\etal~\cite{thies2015real} present expression swap for facial reenactment with an RGB-D camera.
\emph{Face2Face}~\cite{thies2016face2face} is a real-time face reenactment system using only an RGB camera.
Instead of manipulating expression only, the extended work~\cite{kim2018deep} transfers the full $3$D head position, rotation, expression, and eye blinking from a source actor to a portrait video of a target actor.
``Synthesizing Obama"~\cite{suwajanakorn2017synthesizing} animates the face based on an input audio signal.
\emph{FaceSwap} replaces the identity of $3$D models while preserving the expressions.

Deep learning techniques, not surprisingly, are popular in synthesizing or manipulating faces~\cite{on-learning-3d-face-morphable-model-from-in-the-wild-images}.
The term \emph{Deepfakes} has become a synonym for deep learning based face identity replacement~\cite{rossler2019faceforensics++}.
There are various public implementations of \emph{Deepfakes}, most recently by \emph{ZAO}~\cite{Zao} and \emph{FaceAPP}~\cite{FaceApp}.
\emph{FaceAPP} can selectively modify facial attributes~\cite{FaceApp}.
GAN-based methods can produce entire synthetic faces, including non-face background~\cite{disentangled-representation-learning-gan-for-pose-invariant-face-recognition,karras2017progressive,karras2019style}.

\Paragraph{Fake Face Benchmarks} Unfortunately, large and diverse datasets for face manipulation detection are limited in the community.
Zhou~\etal~\cite{zhou2017two} collected a dataset with  face-swapped images generated by an iOS app and an open-source software.
Video-based face manipulation became available with the release of FaceForensics~\cite{rossler2018faceforensics}, which contains $0.5$M \emph{Face2Face} manipulated frames from over $1{,}000$ videos.
An extended version, FaceForensics++~\cite{rossler2019faceforensics++}, further augments the collection with \emph{Deepfake}~\cite{deepfakes} and \emph{FaceSwap} manipulations.
However, these datasets are still limited to two fake types: identity and expression swap.
To overcome this limitation, we collect the first fake face dataset with diverse fake types, including identity and expression swapped images from FaceForensics++, face attribute manipulated images using \emph{FaceAPP}, and complete fake face images using StyleGAN~\cite{karras2019style} and PGGAN~\cite{karras2017progressive}.

\Paragraph{Manipulation Localization}
There are two main approaches to localize manipulated image regions: segmenting the entire image~\cite{badrinarayanan2017segnet, nguyen2019multi}, and repeatedly performing binary classification via a sliding window~\cite{rossler2019faceforensics++}.
These methods are often implemented via multi-task learning with additional supervision, yet they do not necessarily improve the final detection performance.
In contrast, we propose an \emph{attention mechanism} to automatically detect the manipulated region for face images, which requires very few additional trainable parameters.
In computer vision, attention models have been widely used for image classification~\cite{cao2015look, xiao2015application, wang2017residual}, image inpainting~\cite{liu2018image, yu2018generative} and object detection~\cite{yoo2015attentionnet, caicedo2015active}.
Attention not only serves to select a focused location but also enhances object representations at that location, which is effective for learning generalizable features for a given task.
A number of methods~\cite{wang2018non, woo2018cbam, hu2018squeeze} utilize an attention mechanism to enhance the accuracy of CNN classification models.
Residual Attention Network~\cite{wang2017residual} improves the accuracy of the classification model using $3$D self-attention maps.
Choe~\etal~\cite{choe2019attention} propose an attention-based dropout layer to process the feature maps of the model, which  improves the localization accuracy of  CNN classifiers. 
To our knowledge, this is the \emph{first} work to develop the attention mechanism to face manipulation detection and localization.

\section{Proposed Method}
\label{sec:method}

We pose the manipulated face detection as a binary classification problem using a CNN-based network. We further propose to utilize the attention mechanism to process the feature maps of the classifier model.
The learned attention maps can highlight the regions in an image which influence the CNN's decision, and further be used to guide the CNN to discover more discriminative features.

\subsection{Motivation for the Attention Map}

Assuming the attention map can highlight the manipulated image regions, and thereby guide the network to detect these regions, this alone should be useful for the face forgery detection.
In fact, each pixel in the attention map would compute a probability that its receptive field corresponds to a manipulated region in the input image.
Digital forensics has shown that camera model identification is possible due to ``fingerprints'' in the high-frequency information of a real image~\cite{chang_cvpr20}.
It is thus feasible to detect abnormalities in this high-frequency information due to algorithmic processing. 
Hence we insert the attention map into the backbone network where the receptive field corresponds to appropriately sized local patches.
Then, the features before the attention map encode the high-frequency fingerprint of the corresponding patch, which may discriminate between real and manipulated regions at the local level.

Three major factors were considered during the construction and development of our attention map; \textit{i}) explainability, \textit{ii}) usefulness, and \textit{iii}) modularity.

\textbf{Explainability}: Due to the fact that a face image can be modified entirely or in part, we produce an attention map that predicts where the \textit{modified} pixels are.
In this way, an auxiliary output is produced to explain which spatial regions the network based its decision on.
This differs from prior works in that we use the attention map as a mask to remove any irrelevant information from the high-dimensional features \textit{within the network}.
During training, for a face image where the entire image is real, the attention map should ignore the entire image.
For a modified or generated face, at least some parts of the image are manipulated, and therefore the ideal attention map should focus only on these parts.

\textbf{Usefulness}: One objective of our proposed attention map is that it enhances the final binary classification performance of the network.
This is accomplished by feeding the attention map back into the network to ignore non-activated regions.
This follows naturally from the fact that modified images may only be \textit{partially modified}.
Through the attention map, we can remove the real regions of a partial fake image so that the features used for final binary classification are purely from modified regions.

\textbf{Modularity}: To create a truly utilitarian solution, we take great care to maintain the modularity of the solution.
Our proposed attention map can be implemented easily and plugged into existing backbone networks, through the inclusion of a single convolution layer, its associated loss functions, and masking the subsequent high-dimensional features.
This can even be done while leveraging pre-trained networks by initializing only the weights that are used to produce the attention map.

\subsection{Attention-based Layer}

As shown in Fig.~\ref{fig:network_architecture}, the attention-based layer can be applied to any feature map of a classification model, and focus the network's attention on discriminative regions.
Specifically, the input of the attention-based layer is a convolutional feature map $\mathbf{F}\in\mathbb{R}^{H\times W\times C}$, where $H$, $W$, $C$ are height, width, and the number of channels, respectively. For simplicity, we omit the mini-batch dimension in this notation.
Then we can generate an attention map $\mathbf{M}_{att}=\Phi(\mathbf{F})\in\mathbb{R}^{H\times W}$ by processing $\mathbf{F}$, where $\Phi(\cdot)$ denotes the processing operator. The output of attention module is the refined feature map $\mathbf{F}'$, which is calculated as:  
\begin{equation}
\mathbf{F}' = \mathbf{F} \odot \text{Sigmoid}(\mathbf{M}_{{att}}),
    \label{eq:attention}
\end{equation}
where $\odot$ denotes element-wise multiplication. The intensity of each pixel in the attention map is close to $0$ for the real regions, and close to $1$ for the fake regions.
In other words, the pixel of the attention map indicates the probability of the original image patch being a fake region. This helps the subsequent backbone network to focus its processing to the non-zeros areas of the attention map, \emph{i.e.}, the fake regions.
Here, we propose two approaches to implement $\Phi(\cdot)$: manipulation appearance model and direct regression.

\Paragraph{Manipulation Appearance Model (MAM)}
We assume that any manipulated map can be represented as a linear combination of a set of map prototypes:
\begin{equation}
\mathbf{M}_{{att}} = \bar{\mathbf{M}} + \mathbf{A} \cdot \alpha,
    \label{eq:reg_template}
\end{equation}
where $\bar{\mathbf{M}} \in \mathbb{R}^{(H\cdot W)\times 1}$ and $\mathbf{A}\in \mathbb{R}^{(H\cdot W)\times n}$ are the pre-defined average map and basis functions of maps.
Thus the attention map generation can be translated to estimate the weight parameter $\alpha\in \mathbb{R}^{n\times 1}$, for each training image.
We utilize one additional convolution and one fully connected layer to regress the weights from the feature map $\mathbf{F}$ (Fig.~\ref{fig:network_architecture}). 

\begin{figure}[t]
\begin{center}
\includegraphics[width=0.95\linewidth]{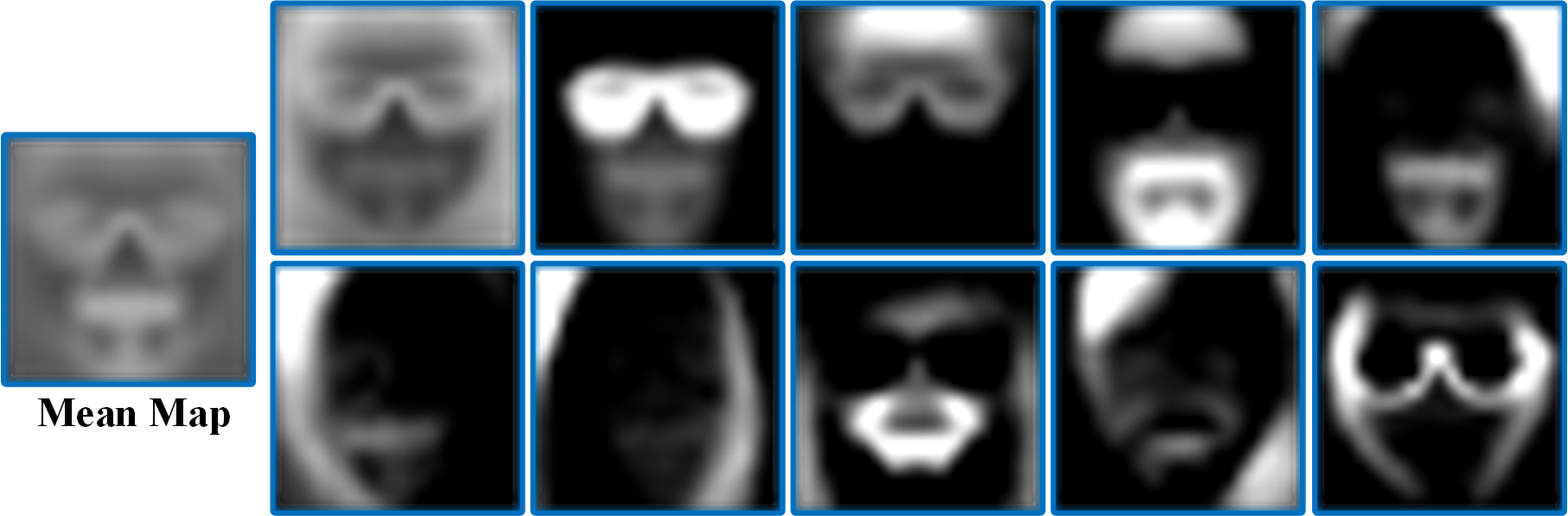}
\end{center}
\figvspace
\vspace{2mm}
\caption{\small Mean map $\bar{\mathbf{M}}$ and $10$ basis components $\mathbf{A}$.}
\label{fig:template}
\figvspace
\end{figure}

The benefit of our proposed MAM is two fold. First, this constrains the solution space of map estimation.
Second, the complexity of the attention estimation is decreased, which is helpful for generalization. To calculate the statistical bases $\mathbf{A}$, we apply Principal Component Analysis (PCA) to $100$ ground-truth manipulation masks computed from \emph{FaceAPP}. The first $10$ principal components are used as bases, \emph{i.e.}, $n=10$. Fig.~\ref{fig:template} shows the mean map and $10$ bases (or templates).

\Paragraph{Direct Regression}
Another way to implement $\Phi(\cdot)$ is to estimate the attention map via a convolutional operation $f$: $\mathbf{F}$ $\overset{f}{\rightarrow}$ $\mathbf{M}_{att}$. $f$ can consist of multiple convolutional layers or a single layer. This direct regression method is simple, yet effective, for adaptive feature refinement.
Later we show that the benefits of our proposed attention-based layer are realized regardless of the choice of backbone networks.
This further validates our claim that the proposed solution is modular and improves the usefulness and flexibility of the attention map.

\subsection{Loss Functions}
\label{sec::los}
To train the binary classification network, we may begin with a pre-trained backbone network or to learn the backbone from scratch.
Either way, the overall training loss is:

\begin{equation}
    \mathcal{L} =  \mathcal{L}_{\text{classifier}} + \lambda * \mathcal{L}_{\text{map}}, 
    \label{eq:network_loss}
\end{equation}
where $ \mathcal{L}_{\text{classifier}}$ is the binary classification loss of Softmax and $\mathcal{L}_{\text{map}}$ is the attention map loss.
$\lambda$ is the  loss weight. 

For attention map learning, we consider three different cases: supervised, weakly supervised, and unsupervised.

\begin{table*}[t!]
    \centering
    \small
    \renewcommand\arraystretch{1}
    \newcommand{\tabincell}[2]{\begin{tabular}{@{}#1@{}}#2\end{tabular}}
    \caption{Comparison of fake face datasets along different aspects: number of still images, number of videos, number of fake types (identity swap (Id.~swap), expression swap (Exp.~swap), attributes manipulation, and entire image synthesis (Entire syn.)) and pose variation.}
    \vspace{1mm}
    \resizebox{0.9\textwidth}{!}{
    \begin{tabular}{c|c|c|c|c|c|c|c|c|c|c}
         \toprule
        \multirow{2}{*}{Dataset} & \multirow{2}{*}{Year} & \multicolumn{2}{|c|}{ \# Still images} & \multicolumn{2}{|c|}{ \# Video clips} & \multicolumn{4}{|c|}{\# Fake types} & {Pose} \\
        \cline{3-10}
       & & Real & Fake & Real & Fake & \tabincell{c}{Id.~swap} & \tabincell{c}{Exp.~swap} & \tabincell{c}{Attr.~mani.} & \tabincell{c}{Entire~syn.} & variation\\ \hline\hline
        Zhou~\etal~\cite{zhou2017two} &$2018$&$2,010$&$2,010$&-&-&$2$&-&-&-&Unknown  \\ \hline
        Yang~\etal~\cite{yang2019exposing} &$2018$&$241$&$252$&$49$&$49$&$1$&-&-&-&Unknown  \\ \hline
        Deepfake~\cite{korshunov2018deepfakes} &$2018$&-&-&-&$620$&$1$&-&-&-&Unknown  \\ \hline    
        FaceForensics++~\cite{rossler2019faceforensics++} &$2019$&-&-&$1,000$&$3,000$&$2$&$1$&-&-& $[-30^{\circ},30^{\circ}]$\\ \hline
        FakeSpotter~\cite{2019arXiv190906122W} & $2019$&$6,000$&$5,000$&-&-&-&-&-&$2$&Unknown\\ \hline
        %\textbf{DFFD} (our) &$2019$&$58,703$&$240,527$&$1,000$&$3,000$&$2$&$1$&$28+40$&$2$&$[-90^{\circ},90^{\circ}]$\\ 
        \textbf{DFFD} (our) &$2019$&$58,703$&$240,336$&$1,000$&$3,000$&$2$&$1$&$28+40$&$2$&$[-90^{\circ},90^{\circ}]$\\ 
        \bottomrule
    \end{tabular}
    }
    \label{tab:data_comp}
    \figvspace
\end{table*}

\Paragraph{Supervised learning}
If the training samples are paired with ground truth attention masks, we can train the network in a supervised fashion, using Eqn.~\ref{eq:supr_loss}.
\begin{equation}
    \mathcal{L}_{\text{map}} =
      ||\mathbf{M}_{{att}}-\mathbf{M}_{{gt}}||_1,\\
    \label{eq:supr_loss}
\end{equation}
where $\mathbf{M}_{gt}$ is the ground truth manipulation mask. 
We use zero-maps as the $\mathbf{M}_{gt}$ for real faces, and one-maps as the $\mathbf{M}_{gt}$ for entirely synthesized fake faces.
For partially manipulated faces, we pair fake images with their corresponding source images, compute the absolute pixel-wise difference in the RGB channels, convert into grayscale, and divide by $255$ to produce a map in the range of $[0,1]$.
We empirically determine the threshold of $0.1$ to obtain the binary modification map as $\mathbf{M}_{gt}$.
We posit this strong supervision can help attention-based layer to learn the most discriminative regions and features for fake face detection.

\Paragraph{Weakly supervised learning}
For partially manipulated faces, sometimes the source images are not available. 
Hence, we can not obtain the ground truth  manipulation mask as described above.
However, we would still like to include these faces in learning the attention maps.
To this end, we propose a weak supervision map loss as in Eqn.~\ref{eq:amap_loss}:
\begin{equation}
    \mathcal{L}_{\text{map}} = 
    \begin{cases}
        |\text{Sigmoid}(\mathbf{M}_{{att}})-0|, & \text{if }  \text{real} \\
        |\max(\text{Sigmoid}(\mathbf{M}_{{att}}))-0.75|. & \text{if } \text{fake} \\
    \end{cases}
    \label{eq:amap_loss}
\end{equation}
%where $\mathbf{M}^i_{att}$ is the attention map of the $i$-th sample in a mini-batch.
This loss drives the attention map to remain un-activated for real images, {\it i.e.}, all $0$.
For fake images, regardless of entire or partial manipulation, the maximum map value across the entire map should be sufficiently large, $0.75$ in our experiments. 
Hence, for partial manipulation, an arbitrary number of the map values can be zeros, as long as at least one modified local region has a large response.

\Paragraph{Unsupervised learning} 
The proposed attention module can also allow us to train the network without any map supervision when $\lambda_m$ is set to $0$.
With only image-level classification supervision, the attention map learns informative regions automatically.
More analysis of these losses is available in the experiments section.

\section{Diverse Fake Face Dataset}\label{sec:dataset}

One of our contributions is the construction of a dataset with diverse types of fake faces, termed Diverse Fake Face Dataset (DFFD). Compared with previous datasets in Tab.~\ref{tab:data_comp}, DFFD contains greater diversity, which is crucial for detection and localization of face manipulations.

\Paragraph{Data Collection} 
In Sec.~\ref{sec:intro}, we introduced four main facial manipulation types: identity swap, expression swap, attribute manipulation, and entire synthesized faces. 
We thus collect data from these four categories by adopting respective state-of-the-art (SOTA) methods to generate fake images.
Among all images and video frames, $47.7\%$ are from male subjects, $52.3\%$ are from females, and the majority of samples are from subjects in the range $21$-$50$ years of age. 
For the face size, both real and fake samples have both low quality and high quality images.
This ensures that the distributions of gender, age, and face size are less biased. 

\emph{Real face images.}
We utilize FFHQ~\cite{karras2019style} and CelebA~\cite{liu2015faceattributes} datasets as our real face samples since the faces contained therein cover comprehensive variations in race, age, gender, pose, illumination, expression, resolution, and camera capture quality.
We further utilize the source frames from FaceForensics++~\cite{rossler2019faceforensics++} as additional real faces.

\emph{Identity and expression swap.} For facial identity and expression swap, we use all the video clips from FaceForensics++~\cite{rossler2019faceforensics++}. 
The FaceForensics++ contains $1{,}000$ real videos collected from YouTube and their corresponding $3{,}000$ manipulated versions which are divided into two groups: identity swap using \emph{FaceSwap} and \emph{Deepfake}~\cite{deepfakes}, and expression swap using \emph{Face2Face}~\cite{thies2016face2face}.
From a public website~\cite{footnote1}, we collect additional identity swap data, which are videos generated by \emph{Deep Face Lab} (DFL)~\cite{footnote2}.

\begin{figure}[t]
\begin{center}
\includegraphics[width=0.98\linewidth]{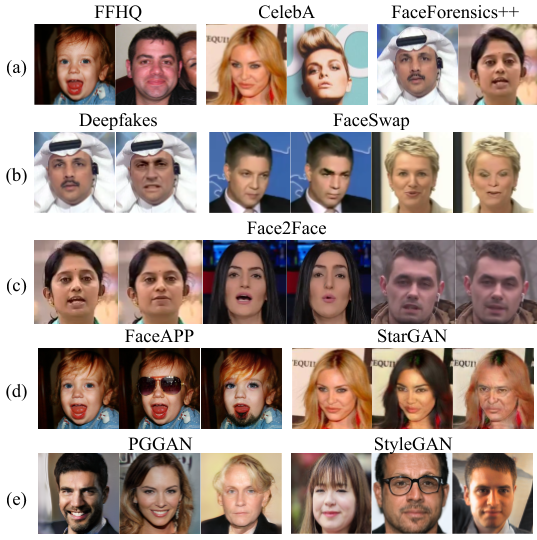}
\end{center}
\figvspace
\vspace{1mm}
\caption{\small Example faces in our DFFD. (a) Real images/frames from FFHQ, CelebA and FaceForensics++ datasets; (b) Paired face identity swap images from FaceForensics++ dataset; (c) Paired face expression swap images from FaceForensics++ dataset; (d) Attributes manipulated examples by \emph{FaceAPP} and StarGAN; (e) Entire synthesized faces by PGGAN and StyleGAN.}
\label{fig:dataset_examples}
\figvspace
\end{figure}

\emph{Attributes manipulation.} 
We adopt two methods~\emph{FaceAPP}~\cite{FaceApp} and StarGAN~\cite{choi2018stargan} to generate attribute manipulated images, where $4{,}000$ faces of FFHQ and $2{,}000$ faces of CelebA are the respective input real images.
~\emph{FaceAPP}, as a consumer-level smartphone app, provides $28$ filters to modify specified facial attributes, \emph{e.g.}, gender, age, hair, beard, and glasses. 
The images are randomly modified with an automated script running on Android devices.
For each face in FFHQ, we generate three corresponding fake images: two with a single random manipulation filter, and one with multiple manipulation filters.
For each face in CelebA, we generate $40$ fake images by StarGAN, a GAN-based image-to-image translation method. In total, we collect $92$K attribute manipulated images.

\emph{Entire face synthesis.} 
Recent works such as PGGAN~\cite{karras2017progressive} and StyleGAN~\cite{karras2019style} achieve remarkable success in realistic face image synthesis.
PGGAN proposes a progressive training scheme both for generator and discriminator, which can produce  high-quality images.
StyleGAN redesigns the generator by borrowing from style transfer literature.
Thus, we use the pre-trained model of PGGAN and StyleGAN to create $200$k and $100$k high-quality entire fake images, respectively.
Figure~\ref{fig:dataset_examples} shows examples of DFFD.

\Paragraph{Pre-processing}
InsightFace~\cite{guo2018stacked} is utilized to estimate the bounding box and $5$ landmarks for each image.
We discard images whose detection or alignment fails.
We further generate ground truth manipulation masks for fake images as described in Sec.~\ref{sec::los}.
To enforce consistency, if a fake face image is derived from a source real face image, we use the same landmarks of the real face image for face cropping.

\Paragraph{Protocols}
We collect $781{,}727$ samples for real image, and $1{,}872{,}007$ samples for fake ones.
Within these samples, we randomly select a subset of $58{,}703$ real images and $240{,}336$ fake ones to make the size of our dataset manageable and to balance the size of each sub-category. 
For video samples, we extract one frame per second in order to reduce the size without sacrificing the diversity of DFFD. 
We randomly split the data into $50\%$ for training, $5\%$ for validation and $45\%$ for testing.
All fake images manipulated from the same real image are in the same set as the source image.

\section{Experimental Results}

%We study three aspects of our proposed method: ablation study, forgery detection, and manipulation localization.

\subsection{Experimental Setup}
\textbf{Implementation Details:} 
The loss weight $\lambda$ is set to $1$ and the batch size is $16$, where each mini-batch consists of $8$ real and $8$ fake images. We use XceptionNet~\cite{chollet2017xception} and VGG16~\cite{simonyan2014very} as backbone networks. Both two networks were pre-trained on ImageNet and fine-tuned on DFFD. Adam optimizer is used with a learning rate of $0.0002$ in all experiments. 
Depending on the backbone architecture, we train for $75$k-$150$k iterations, which requires less than $8$ hours on an NVidia GTX 1080Ti.

\textbf{Metrics:}
For all experiments, we utilize the protocols defined in Sec.~\ref{sec:dataset}.
For detection, we report  Equal Error Rate (EER),  Area Under Curve (AUC) of ROC,  True Detect Rate (TDR) at  False Detect Rate (FDR) of $0.01\%$ (denoted as TDR$_{0.01\%}$), and TDR at FDR of $0.1\%$ (denoted as TDR$_{0.1\%}$). 
For localization, with known ground-truth masks, we report Pixel-wise Binary Classification Accuracy (PBCA), which treats each pixel as an independent sample to measure classification accuracy, Intersection over Union (IoU), and Cosine similarity between two vectorized maps. 
We also propose a novel metric, termed Inverse Intersection Non-Containment (IINC) for evaluating face manipulation localization performance, as described in Sec.~\ref{sec:map}.

\subsection{Ablation Study}

\textbf{Benefit of Attention map:}
We utilize the SOTA XceptionNet~\cite{chollet2017xception}  as our backbone network.
It is based on depth-wise separable convolution layers with residual connections.
We convert XceptionNet into our model by inserting the attention-based layer between Block $4$ and Block $5$ of the middle flow, and then fine-tune on DFFD training set.

\begin{table}[t]
\renewcommand\arraystretch{1}
    \centering
    \caption{Ablation for the benefit of the attention map, with various combinations of map generation methods and supervisions. [Key: top performance for \textbf{supervised} and \textcolor{blue}{ weakly supervised} methods]
    }
    \vspace{1mm}
    \resizebox{\linewidth}{!}{
    \begin{tabular}{l|c|c|c|c|c}
        \toprule
        Map Supervision & AUC & EER & TDR$_{0.01\%}$ & TDR$_{0.1\%}$ & PBCA \\ \hline \hline
        Xception & $99.61$ & $2.88$ & $77.42$ & $85.26$ & $-$ \\ \hline
        + Reg., \emph{unsup.} & $99.76$ & $2.16$ & $77.07$ & $89.70$ & $12.89$ \\ \hline
        + Reg., \emph{weak sup.}  & $99.66$ & $\color{blue}{2.57}$ & $46.57$ & $75.20$ & $30.99$ \\ \hline
        + Reg., \emph{sup.} & $99.64$ & $\textbf{2.23}$ & $\textbf{83.83}$ & $\textbf{90.78}$ & $\textbf{88.44}$ \\ \hline
        + Reg., \emph{sup.} - map & $\textbf{99.69}$ & ${2.73}$ & ${48.54}$ & ${72.94}$ & $\textbf{88.44}$ \\  \hline  
        + MAM, \emph{unsup.} & $99.55$ & $3.01$ & $58.55$ & $77.95$ & $36.66$ \\ \hline
        + MAM, \emph{weak sup.}  & $\color{blue}{99.68}$ & $2.64$ & $\color{blue}{72.47}$ & $\color{blue}{82.74}$ & $\color{blue}{69.49}$  \\ \hline
        + MAM, \emph{sup.} & $99.26$ & $3.80$ & $77.72$ & $86.43$ & $85.93$  \\ \hline
        + MAM, \emph{sup.} - map & $98.75$ & $6.24$ & $58.25$ & $70.34$ & $85.93$  \\  \bottomrule 
    \end{tabular}
    }
    \label{tab:ablation_study}
    \figvspace
\end{table}

In Tab.~\ref{tab:ablation_study}, we show a comparison of the direct regression (Reg.)~and manipulation appearance model (MAM) with different supervision strategies, {\it i.e.}, unsupervised (\emph{unsup.}), weakly supervised (\emph{weak sup.}) and supervised (\emph{sup.}) learning.
While four detection metrics are listed for completeness, considering the overall strong performance of some metrics and the preferred operational point of low FDR in practice, TDR at low FDR (\emph{i.e.,} TDR$_{0.01\%}$) should be the primary metric for comparing various methods.

Unsurprisingly, the supervised learning outperforms the weakly supervised and unsupervised, in both the detection and localization accuracies.
Additionally, comparing two map estimation approaches, regression-based method performs better with supervision. In contrast, MAM-based method is superior for weakly supervised or  unsupervised cases as MAM offers strong constraint for map estimation. 

Finally, instead of using the softmax output, an alternative is to use the average of the estimated attention map for detection, since loss functions encourage low attention values
for real faces while higher values for fake ones.
The performance of this alternative is shown in the rows of `*, \emph{sup.} - map' in Tab.~\ref{tab:ablation_study}.
Although this is not superior to the softmax output, it demonstrates that the attention map is itself useful for the facial forgery detection task.

\begin{table}[t]
\renewcommand\arraystretch{1}
    \centering
    \caption{Our attention layer in two  backbone networks.}
    \vspace{1mm}
    \resizebox{0.98\linewidth}{!}{
    \begin{tabular}{l|c|c|c|c|c}
        \toprule
        Network & AUC & EER & TDR$_{0.01\%}$ & TDR$_{0.1\%}$ & PBCA\\ \hline \hline
        Xception & $99.61$ & $2.88$ & $77.42$ & $85.26$ & -\\
        Xception + Reg.  & $\textbf{99.64}$ & $\textbf{2.23}$ & $\textbf{83.83}$ & $\textbf{90.78}$  & $\textbf{88.44}$\\
        Xception + MAM   & $99.26$ & $3.80$ & $77.72$ & $86.43$ & $85.93$ \\ \hline
        VGG16 & $96.95$ & $8.43$ & $0.00$ & $51.14$ & -\\
        VGG16 + Reg.  & $99.46$ & $3.40$ & $44.16$ & $61.97$ & $\textbf{91.29}$\\ 
        VGG16 + MAM & $\textbf{99.67}$ & $\textbf{2.66}$ &  $\textbf{75.89}$ & $\textbf{87.25}$ & $86.74$\\ \hline
    \end{tabular}
    }
    \label{tab:backbone_validation}
    %\figvspace
    \vspace{1mm}
\end{table}

\begin{figure}[t]
\begin{center}
\includegraphics[width=0.85\linewidth]{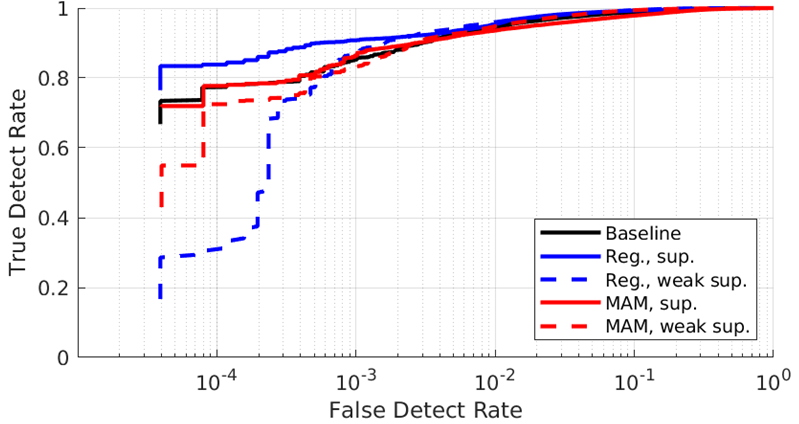}
\end{center}
\figvspace
\vspace{2mm}
\caption{\small Forgery detection ROCs of the XceptionNet backbone with and without the attention mechanism.}
\label{fig:map_type_rocs}
\figvspace
\vspace{1mm}
\end{figure}

\textbf{Effect on Backbone Networks:}
We also report the results of a shallower backbone network VGG$16$~\cite{simonyan2014very}. Tab.~\ref{tab:backbone_validation} compares XceptionNet and VGG$16$ with and without the attention layer. Both Reg.~and MAM models are trained in the supervised case. We observe that using attention mechanism does improve the detection on both backbones.

Specifically, with a large and deep network (XceptionNet), the attention map can be directly produced by the network given the large parameter space.
This directly produced attention map can better predict the manipulated regions than a map estimated from the MAM bases.
Inversely, when using a smaller and shallower network (VGG16), we find that the direct production of the attention map causes contention in the parameter space.
Hence including the prior of the MAM bases reduces this contention and allows for increased detection performance, though its estimation of the manipulated regions is constrained by the map bases.

\subsection{Forgery Detection Results}
\begin{figure}[t!]
\begin{center}
\includegraphics[width=0.95\linewidth]{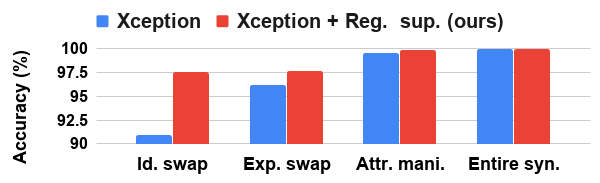}
\end{center}
\vspace{-4mm}
\caption{\small Binary classification accuracy for different fake types.}
\label{fig:sample1}
\vspace{-1mm}
\end{figure}

We first show the ROCs on our DFFD in Fig.~\ref{fig:map_type_rocs}.
Obviously, the direct regression approach for the attention map produces the best performing network at low FDR, which is not only the most challenging scenario, but also the most relevant to the practical applications. In addition, the proposed attention layer substantially outperforms the conventional XceptionNet, especially at lower FDR.
Fig.~\ref{fig:sample1} plots binary classification accuracy of our \emph{Reg., sup} and baseline for different fake types of DFFD. The proposed approach benefits forgery detection of all considered fake types, especially for the facial identity and expression swap.

\begin{figure*}[t!]
\begin{center}
\includegraphics[width=\linewidth]{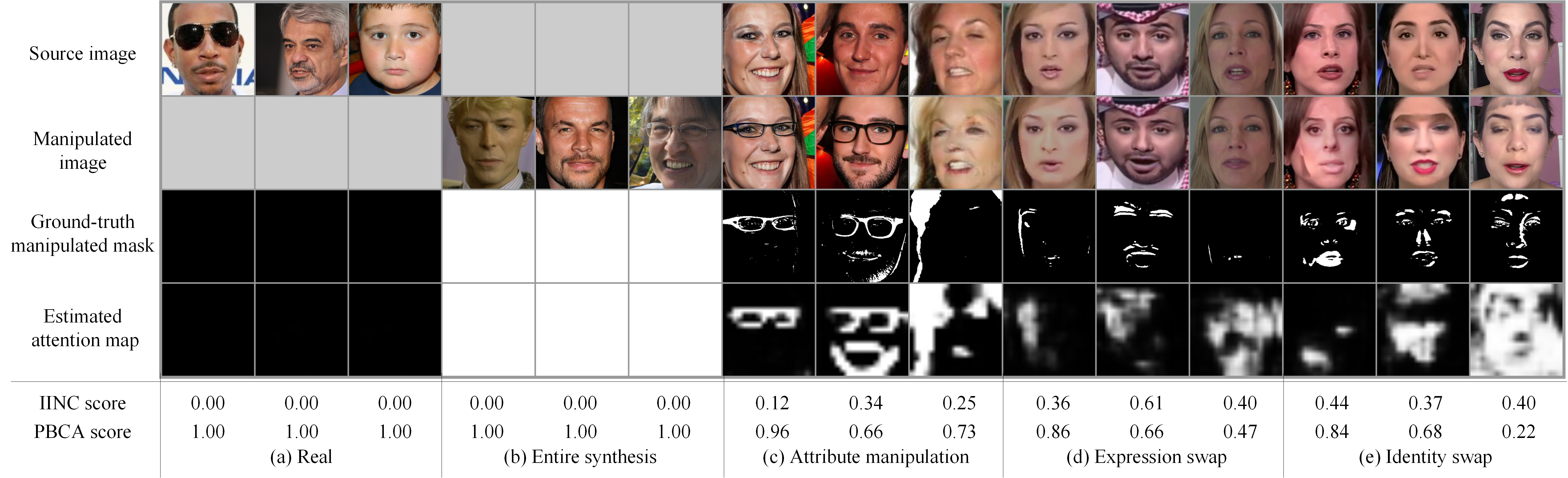}
\end{center}
\vspace{-2mm}
\caption{\small Estimated attention maps by applying Xception + Reg.~\textit{sup.} model to real and $4$ types of manipulated images, with IINC and PBCA scores computed w.r.t.~ground truth. While the overall areas of the attention maps are correct, their fidelity could be further improved.}
\label{fig:map_pred}
\vspace{-3mm}
\end{figure*}

We further validate our model on public datasets, where SOTA facial forgery detection methods have been tested.
Table~\ref{tab:baselines_comparison} summarizes the performance of all methods. % on these datasets.
Note that the performances shown here are not strictly comparable since not all methods are trained on the same dataset. 
First, we evaluate on the UADFV and Celeb-DF datasets with the models trained with DFFD. As shown in Tab.~\ref{tab:baselines_comparison}, our proposed approach significantly outperforms all the baselines on Celeb-DF and achieves competitive results on UADFV. FWA~\cite{li2018exposing} and HeadPose~\cite{yang2019exposing} demonstrate superior performance on UADFV partially because they are trained on the same UADFV dataset, while this data source is not in our DFFD. 
Second, for a fair comparison, we train our method and baseline Xception on UADFV training set.
In this case, our method outperforms all baselines on UADFV, and still shows superior generalization on Celeb-DF.
Third, the results in Tab.~\ref{tab:baselines_comparison} also help us to identify both the \emph{source} and \emph{amount} of improvements.
For example, $75.6\%\rightarrow84.2\%$ is an improvement due to attention mechanism while
$52.2\%\rightarrow63.9\%$ and $57.1\%\rightarrow64.4\%$ are due to the greater diversity of DFFD dataset.

\begin{table}[t]
\renewcommand\arraystretch{0.98}
\footnotesize
    \centering
    \caption{\small AUC (\%) on UADFV and Celeb-DF. All baseline results are quoted from~\cite{li2019celeb}}
    \vspace{1mm}
    \resizebox{0.99\linewidth}{!}{
    \begin{tabular}{l|c|c|c}
        \toprule
        Methods & Training data & UADFV~\cite{yang2019exposing} & Celeb-DF~\cite{li2019celeb} \\
        \hline \hline
        Two-stream~\cite{zhou2017two} & Private data & $85.1$ & $53.8$ \\\hline
        Meso4~\cite{afchar2018mesonet} & \multirow{2}{*}{Private data} & $84.3$ & $54.8$ \\
        MesoInception4~\cite{afchar2018mesonet} & & $82.1$ & $53.6$ \\\hline
        HeadPose~\cite{yang2019exposing}& UADFV & $89.0$ & $54.6$ \\\hline
        FWA~\cite{li2018exposing}& UADFV & $97.4$ & $56.9$ \\\hline
        VA-MLP~\cite{matern2019exploiting} & \multirow{2}{*}{Private data} & $70.2$ & $55.0$ \\
        VA-LogReg~\cite{matern2019exploiting} &  & $54.0$ & $55.1$ \\\hline
        Multi-task~\cite{nguyen2019multi} & FF & $65.8$ & $54.3$ \\\hline
        Xception-FF++~\cite{rossler2019faceforensics++} & FF++ & $80.4$ & $48.2$ \\\hline \hline
        Xception & DFFD & $75.6$ & $63.9$ \\\hline
        Xception & UADFV & $96.8$ & $52.2$ \\\hline
        Xception & UADFV, DFFD & $97.5$ & $67.6$ \\\hline\hline
      Xception+Reg.& DFFD & $84.2$ & $64.4$ \\ \hline
        Xception+Reg. & UADFV & $\mathbf{98.4}$ & $57.1$ \\ \hline
        Xception+Reg. & UADFV, DFFD & $\mathbf{98.4}$ & $\mathbf{71.2}$ \\ \toprule
    \end{tabular}
    }
    \label{tab:baselines_comparison}
    \figvspace
\end{table}

\subsection{Manipulation Localization Results} \label{sec:map}

We utilize three metrics for evaluating the attention maps: Intersection over Union (IoU), Cosine Similarity, and Pixel-wise Binary Classification Accuracy (PBCA).
However, these three metrics are inadequate for robust evaluation of these diverse maps.
Thus, we propose a novel metric defined in Eqn.~\ref{eq:map_metric}, termed Inverse Intersection Non-Containment (IINC), to evaluate the predicted maps:
\begin{equation}
\footnotesize
    \text{IINC} = \frac{1}{3-\mathbf{|U|}} *
    \begin{cases}
        0 & \hspace{-5mm} \text{if } \overline{\mathbf{M}_{\text{gt}}} = 0 \text{ and } \overline{\mathbf{M}_{\text{att}}} = 0 \\
        1 & \hspace{-5mm} \text{if } \overline{\mathbf{M}_{\text{gt}}} = 0 \text{ xor } \overline{\mathbf{M}_{\text{att}}} = 0 \\
        (2-\frac{\mathbf{|I|}}{|\mathbf{M}_{\text{att}}|}-\frac{\mathbf{|I|}}{|\mathbf{M}_{\text{gt}}|}) & \text{otherwise}, \\
    \end{cases}
    \label{eq:map_metric}
\end{equation}
where $\mathbf{I}$ and $\mathbf{U}$ are the intersection and union between the ground truth map, $\mathbf{M}_{gt}$, and the predicted map, $\mathbf{M}_{att}$, respectively.  $\overline{\mathbf{M}}$ and $|\mathbf{M}|$ are the mean and $L_1$ norm of $\mathbf{M}$, respectively.
The two fractional terms measure the ratio of the area of the intersection w.r.t.~the area of each map, respectively.
IINC improves upon other metrics by measuring the non-overlap ratio of both maps, rather than their combined overlap, as in IoU.
Additionally, the IoU and Cosine Similarity are undefined when either map is uniformly $0$, which is the case for real face images.

\begin{table}
	\centering
	\small
	\caption{\small Evaluating manipulation localization with $4$ metrics. }
	\vspace{1mm}
	\resizebox{0.85\linewidth}{!}{
		\begin{tabular}{l|c|c|c|c}
			\toprule
			Data     & IINC $\downarrow$   & IoU $\uparrow$ & Cosine Similarity $\downarrow$ & PBCA $\uparrow$\\ \hline\hline
			All Real & $0.015$  & $-$     & $-$         & $0.998$ \\ \hline
			All Fake & $0.147$ & $0.715$  & $0.192$      & $0.828$ \\ \hline
			Partial  & $0.311$ & $0.401$  & $0.429$      & $0.786$ \\ \hline
			Complete & $0.077$  & $0.847$  & $0.095$       & $0.847$ \\ \hline
			All      & $0.126$ & $-$     & $-$         & $0.855$ \\ \bottomrule
		\end{tabular}
	}
	\label{tab:attention_map_evaluation}
	\figvspace
	\vspace{1mm}
\end{table}

\begin{figure}[t!]
    \centering
    \resizebox{0.95\linewidth}{!}{
    \includegraphics[width=\linewidth]{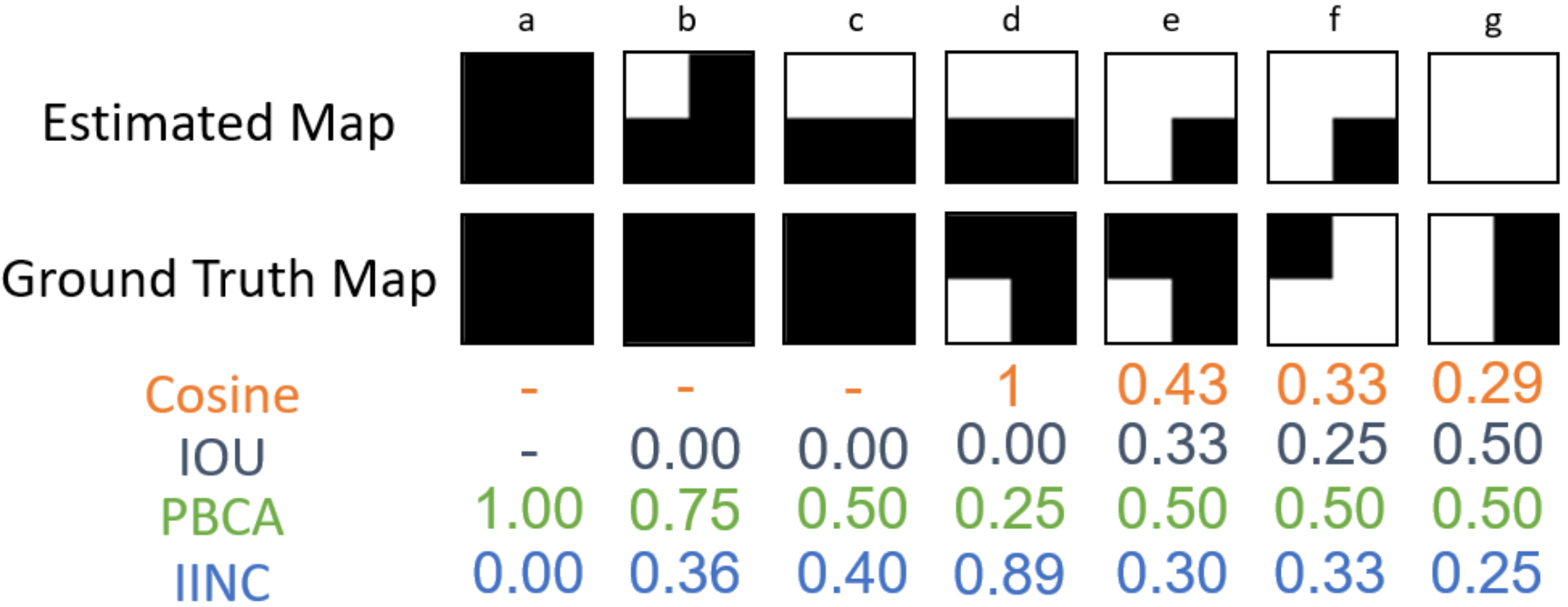}
    }
    \vspace{1mm}
    \caption{A toy-example comparing $4$ metrics in evaluating attention maps. White is the manipulated pixel and black is the real pixel. IOU and Cosine metrics do not adequately reflect the differences in cases (a-c), while PBCA is not useful for cases (e-g). In contrast, the proposed IINC is discriminative in all cases.}
    \label{fig:map_metric}
    \vspace{-3mm}
\end{figure}

The benefits of IINC as compared to other metrics are shown in Fig.~\ref{fig:map_metric}.
Note that IOU and Cosine similarity are not useful for cases (a-c), where the scores are the same, but the maps have vastly different properties.
Similarly, PBCA is not useful for the cases (e-g), as the ratio of mis-classification is not represented in PBCA.
For example, case (g) over-estimates by a factor of $100\%$ and case (e) over-estimates by $200\%$, while case (f) both over- and under-estimates by $150\%$.
The IINC provides the optimal ordering by producing the same order as IOU when it is useful, cases (d-g), and similarly with PBCA when it is useful, cases (a-c).
Thus, IINC is a more robust metric for comparing the attention maps than the previous metrics.

The localization ability of our Xception + Reg.~\textit{sup.} model to predict the attention maps is shown in  Tab.~\ref{tab:attention_map_evaluation}.
In Fig.~\ref{fig:map_pred}, we show the IINC and PBCA for some test examples.
The ordering of the IINC scores aligns with qualitative human analysis.
The first cases in (d) and (e) are examples where the PBCA is high only because the majority of each map is non-activated.
The IINC is more discriminative in these cases due to the non-overlap between the maps.
For the third cases in (d) and (e), the IINC produces the same score because the maps display the same behavior (a large amount of over-activation), whereas the PBCA prefers the example in (d) because its maps have fewer activations.

\section{Conclusion}

We tackle the digitally manipulated face image detection and localization task.
Our proposed method leverages an attention mechanism to process the feature maps of the detection model.
The learned attention maps highlight the informative regions for improving the detection ability and also highlight the manipulated facial regions.
In addition, we collect the first facial forgery dataset that contains diverse types of fake faces.
Finally, we empirically show that the use of our attention mechanism improves facial forgery detection and manipulated facial region localization.
This is the first unified approach that tackles a diverse set of face manipulation attacks, and also achieves the SOTA performance in comparison to previous solutions.

%-------------------------------------------------------------------------

\section*{Supplementary}
\renewcommand{\thesection}{\Alph{section}}
\setcounter{section}{0}
\vspace{-2mm}
In this supplementary material, we provide some details and additional experimental results.
\vspace{-2mm}
\section{Details}
\vspace{-2mm}
Here we will further detail the Diverse Fake Face Dataset and proposed attention map, and analyze additional experiments.

\begin{table*}[t!]
    \centering
    \footnotesize
    \caption{Statistics of our DFFD composition and protocol.}
    %The total number of available samples, training samples, validation samples, and testing samples are $2{,}653{,}734$, $124{,}731$, $12{,}738$ and $161{,}761$, respectively.}
    \vspace{2mm}
    \resizebox{\linewidth}{!}{
    \begin{tabular}{|c|l|c|c|c|c|c|c|}
    \toprule
        \multicolumn{3}{|c|}{Dataset} & \# Total Samples & \# Training & \# Validation & \# Testing & Average face width (pixel)\\ \hline\hline
        \multirow{3}{*}{Real}
        & \multicolumn{2}{|c|}{FFHQ~\cite{karras2019style}} & $70,000$ & $10,000$ & $999$ & $9,000$ & $750$ \\ \cline{2-8}
        & \multicolumn{2}{|c|}{CelebA~\cite{liu2015faceattributes}} & $202,599$ & $9,974$ & $997$ & $8,979$  & $200$ \\ \cline{2-8}
        & \multicolumn{2}{|c|}{Original @ FaceForensics++~\cite{rossler2019faceforensics++}} & $509,128$ & $10,230$ & $998$ & $7,526$ & $200$ \\ \hline
        \multirow{7}{*}{Fake}
        & \multirow{3}{*}{Id. Swap} 
        & DFL & $49,235$ & $10,006$ & $1,007$ & $38,222$ & $200$  \\ \cline{3-8}
        && Deepfakes @ FaceForensics++~\cite{rossler2019faceforensics++} & $509,128$ & $10,230$ & $999$ & $7,517$ & $200$ \\ \cline{3-8}
        && FaceSwap @ FaceForensics++~\cite{rossler2019faceforensics++} & $406,140$ & $8,123$ & $770$ & $6,056$ & $200$ \\ \cline{2-8}
        & \multirow{1}{*}{Exp. Swap} 
        & Face2Face @ FaceForensics++~\cite{rossler2019faceforensics++} & $509,128$ & $10,212$ & $970$ & $7,554$ & $200$  \\ \cline{2-8}
        & \multirow{2}{*}{Attr. Manip.} 
        & FaceAPP~\cite{FaceApp} & $18,416$ & $6,000$ & $1,000$ & $5,000$  & $700$ \\ \cline{3-8}
        && StarGAN~\cite{choi2018stargan} & $79,960$ & $10,000$ & $1,000$ & $35,960$ & $150$ \\ \cline{2-8}
        & \multirow{2}{*}{Entire Syn.} 
        & PGGAN~\cite{karras2017progressive} & $200,000$ & $19,957$ & $1,998$ & $17,950$ & $750$ \\ \cline{3-8}
        && StyleGAN~\cite{karras2019style} & $100,000$ & $19,999$ & $2,000$ & $17,997$ & $750$ \\

     \bottomrule
    \end{tabular}
    }
    \label{tab:data_stt}
    \vspace{-3mm}
\end{table*}

\subsection{DFFD Dataset Details}
\begin{figure}[t]
\includegraphics[trim=45 0 50 0, clip, width=1\linewidth]{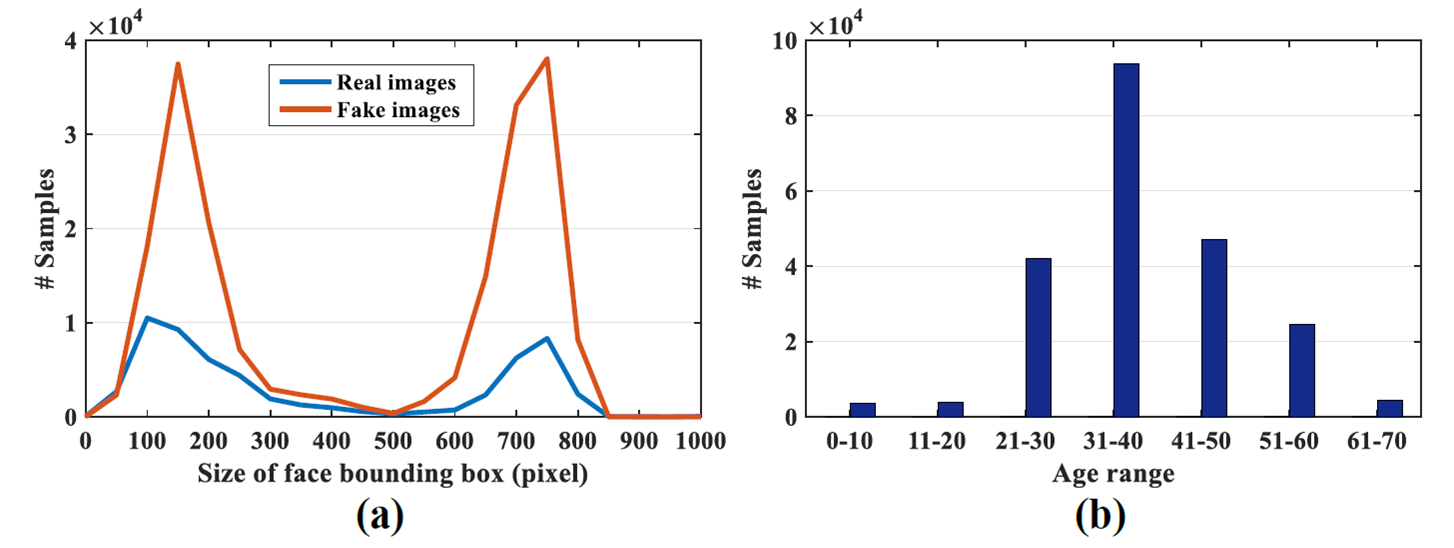}
\vspace{-4mm}
\caption{\small (a) Distribution of the face bounding box sizes (pixel) and (b) Age distribution of our DFFD.}
\label{fig:data_ga_stt}
\vspace{-2mm}
\end{figure}

The DFFD was constructed from large and commonly used facial recognition datasets.
This widespread use of FFHQ and CelebA validate our decision to utilize these as our real images, and the generation of manipulated images from them.
As shown in Fig.~\ref{fig:data_ga_stt}, the DFFD encompasses large variance in both face size and human age, for both real and manipulated images.
Details about the images from datasets used to construct the DFFD are available in Tab.~\ref{tab:data_stt}.

\subsection{Network Architecture Details}
\begin{figure}[t]
\begin{center}
\includegraphics[width=\linewidth]{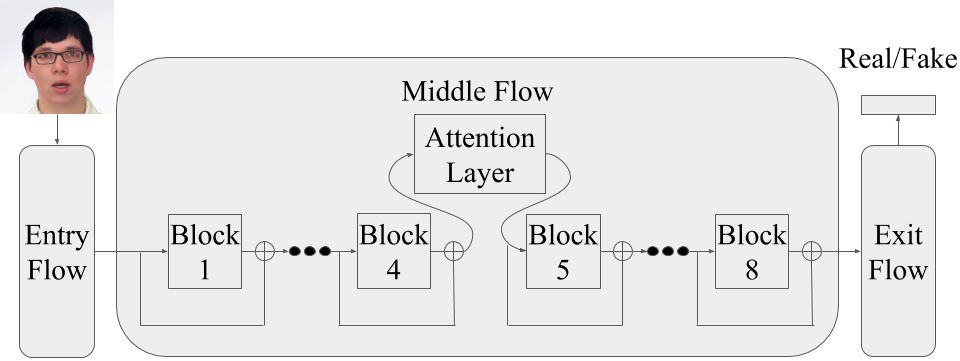}
\vspace{-4mm}
\end{center}
\caption{\small The overall architecture of XceptionNet and its enhancement with our proposed attention later. The original XceptionNet has entry flow, middle flow, and exit flow, where the middle flow is composed of $8$ blocks. Our attention layer can be added after any of the blocks.}
\label{fig:XceptionNet_architecture}
\vspace{-2mm}
\end{figure}

In Fig.~\ref{fig:XceptionNet_architecture}, we show a simplified diagram of the placement of the attention layer within the Xception network.
Due to its modularity, the attention layer can easily be added to any network, in a similar fashion to placing the attention layer in a different location in the Xception network.

\vspace{-2mm}
\section{Additional Experimental Results}

\subsection{Human Study}
We conduct a human study to determine the ability of humans to distinguish between real and manipulated images in the DFFD.
$10$ humans participated in the study.
This was accomplished using a random set of $110$ images from the DFFD, where $10$ images were taken from each row in Tab~\ref{tab:data_stt}.
For each image, the human was required to classify between Real, Entire Fake, and Partial Fake, and additionally required to provide polygon-based regions of interest (attention maps) for Partial Fakes.
The results of this study are shown in Tab.~\ref{tab:human_study}.
It is clear that humans have significant difficulty in the binary classification task (Entire Fake and Partial Fake are considered a single class), while our attention based solution performs almost perfectly.

\begin{figure}[t]
    \centering
    \resizebox{\linewidth}{!}{
    \includegraphics[width=\linewidth]{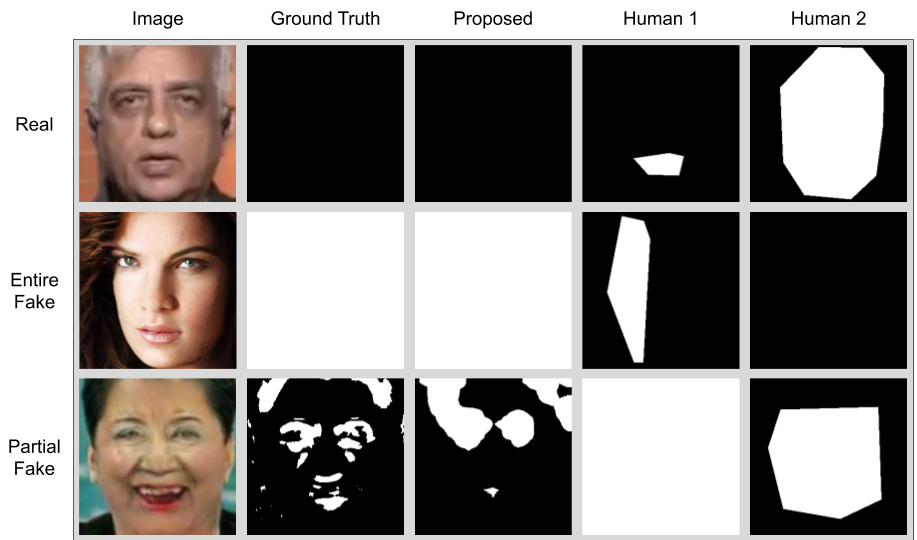}
    }
    \caption{The attention maps produced by our proposed solution and humans during the human study. }
    \label{fig:human_study}
    \vspace{-3mm}
\end{figure}

In Fig.~\ref{fig:human_study}, we show the manipulation maps produced by our proposed solution compared to the maps produced by humans.
Humans focus largely on semantic concepts such as image quality, large artifacts, or strange lighting/color when judging between real and fake images.
Due to this, humans do not detect the very subtle difference in the image ``fingerprint'', which our proposed solution is able to detect.

\begin{table}[t]
\renewcommand\arraystretch{1}
    \centering
        \caption{Comparison between the proposed solution and humans for detecting manipulated images and localization of the manipulated regions. Larger values are better for all but the EER.}
        \vspace{2mm}
             \resizebox{\linewidth}{!}{%
    \begin{tabular}{|c|c|c|c|c|c|c|}
        \hline
        Method & ACC & AUC & EER & TDR$_{0.01\%}$ & TDR$_{0.1\%}$ & PBCA \\ \hline
        Human & $68.18$ & $81.71$ & $30.00$ & $42.50$ & $42.50$ & $58.20$ \\ \hline
        XceptionRegSup & $\textbf{97.27}$ & $\textbf{99.29}$ & $\textbf{3.75}$ & $\textbf{85.00}$ & $\textbf{85.00}$ & $\textbf{90.93}$ \\ \hline
    \end{tabular}
    }
    \label{tab:human_study}
    \vspace{-2mm}
\end{table}

\subsection{Additional Performance Evaluation}
\begin{figure}[t]
    \centering
    \resizebox{\linewidth}{!}{
    \includegraphics[width=\linewidth]{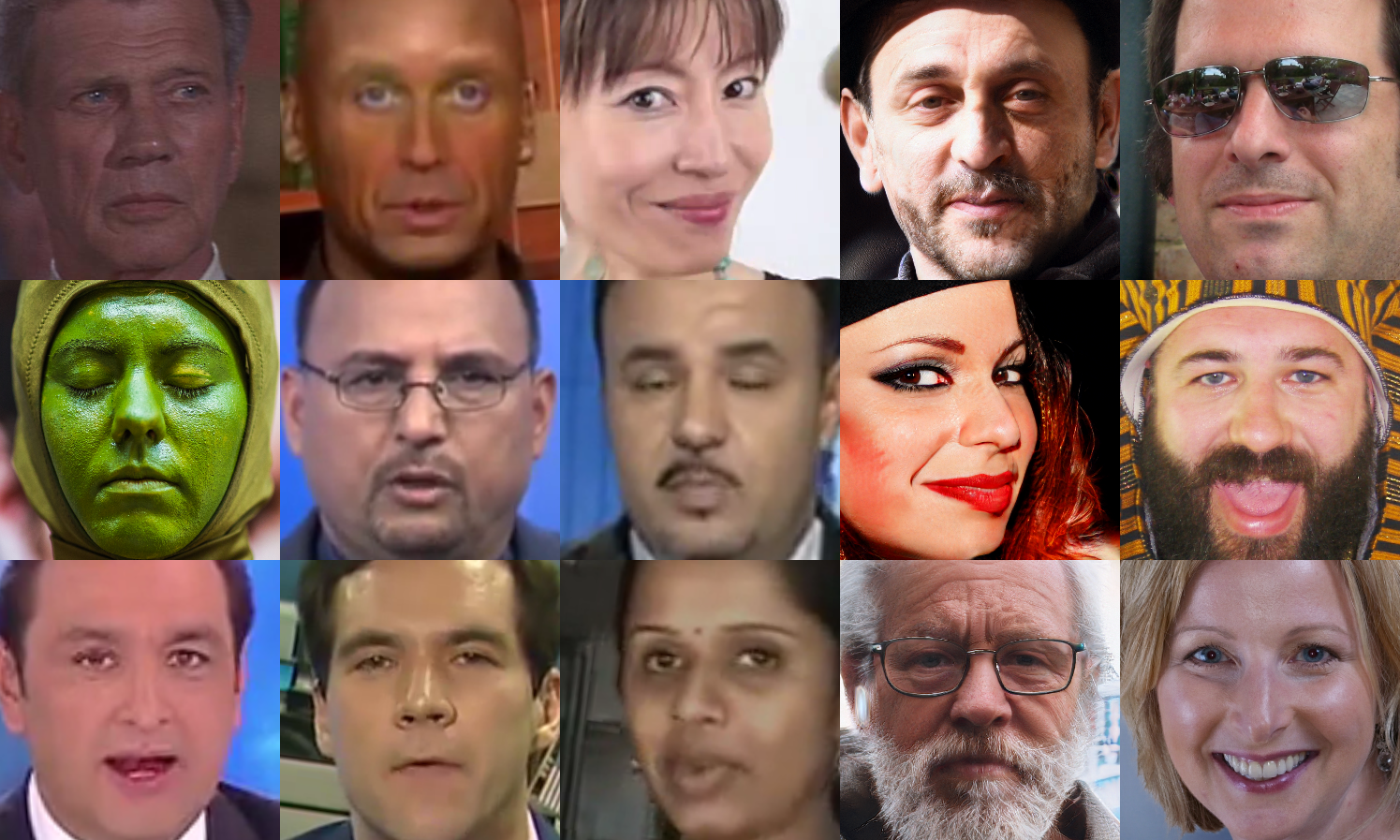}
    }
    \caption{\small Failure examples of the Xception with Regression Map under supervision. 
    From left to right, the columns are top $3$ worst samples of real, identity manipulated, expression manipulated, completely generated, and attribute modified, respectively.}    \label{fig:failures}
    \vspace{-3mm}
\end{figure}

For our best performing model, Xception Regression Map with supervision, we conduct analysis in two aspects.
\textit{i}) Fig.~\ref{fig:failures} shows the worst $3$ test samples among the real test faces and each fake types.
For example, the images in the first column have the lowest Softmax probability of being the real class.
Among these samples, some have heavy makeup, and others are of low image quality.
Meanwhile, the failure cases for the manipulated or entirely synthetic images are high quality and devoid of defects or artifacts.
\textit{ii}) Tab.~\ref{tab:perf_by_type} shows the accuracy of testing samples in each fake type.
The completely synthesized images are the easiest to detect.
This is due to the artificial ``fingerprint'' these methods leave on the generated images, which is easily distinguishable from real images.
In contrast, identity and expression manipulated images are the most challenging to detect, where image is of good quality and no noticeable artifacts exist, as in the $2^{\text{nd}}$ and $3^{\text{rd}}$ columns in  Fig.~\ref{fig:failures}.

\begin{table}[t]
    \centering
    \caption{Fake face detection performance of the Xception Regression Map with supervision for each fake type.}
    \vspace{2mm}
    \resizebox{0.85\linewidth}{!}{
    \begin{tabular}{|c|c|c|c|c|c|}
        \hline
        Fake Type & AUC & EER & TDR$_{0.01\%}$ & TDR$_{0.1\%}$ \\ \hline
        ID Manip. & $99.43$ & $3.11$ & $65.16$ & $77.76$ \\ \hline
        EXP Manip. & $99.40$ & $3.40$ & $71.23$ & $80.87$ \\ \hline
        Attr. Manip. & $99.92$ & $1.09$ & $81.32$ & $90.93$ \\ \hline
        Entire Syn. & $100.00$ & $0.05$ & $99.89$ & $99.96$ \\ \hline
    \end{tabular}
    }
    \label{tab:perf_by_type}
    \vspace{-3mm}
\end{table}

\subsection{Additional Ablation Study}
In Fig~\ref{tab:map_position}, we show an ablation for the placement of the attention layer within the middle flow of the Xception network.
Two trends emerge from this; \textit{i}) the AUC and EER decrease as the attention layer is placed later in the network, and \textit{ii}) the PBCA increases as the attention layer is placed later in the network.
This second trend is expected, the network is able to produce a more finely-tuned attention map given more computational flexibility and depth.
The first trend is more intriguing, because it shows that earlier focus from the attention map is more beneficial for the network than a finely-tuned attention map later.
This earlier attention provides the network with additional time to inspect the features selected by the attention map in order to distinguish between real and manipulated images at a semantic level.

\begin{figure}[t]
\begin{center}
\resizebox{0.99\linewidth}{!}{
\begin{tabular}{@{}c@{}c@{}}
    \includegraphics[trim=20 0 42 27, clip,width=0.8\linewidth]{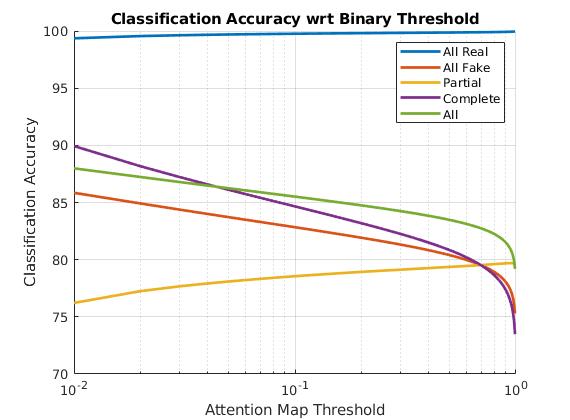} &
    \includegraphics[trim=20 0 42 27, clip,width=0.8\linewidth]{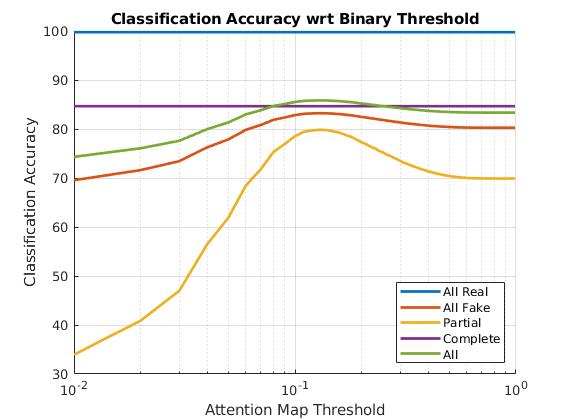} \\
    (a) & (b) \\
\end{tabular}
}
\vspace{-4mm}
\end{center}
\caption{\small The attention map estimation performance of the proposed method when using different thresholds to binarize the predicted map (a) and the ground truth map (b). The threshold for the other map in either case was $0.1$.}
\label{fig:map_thresh_exp}
%\vspace{-5mm}
\end{figure}

In Fig~\ref{fig:map_thresh_exp}, we show the empirical decision for the threshold of $0.1$ that we used to convert maps from continuous values (in the range [$0$,$1$]) to binary values.
This provides strong performance in both graphs of Fig.~\ref{fig:map_thresh_exp}, while being semantically reasonable.
A modification of $0.1$ corresponds to a modification of magnitude equal to $25$ in the typical RGB range of [$0$,$255$].
While a modification of small magnitude ($<10$) is almost undetectable by a human, a modification of larger magnitude ($>25$) is significant.
Therefore, all experiments presented utilized this empirical threshold value of $0.1$.

\begin{table}[t]
\renewcommand\arraystretch{1}
    \centering
        \caption{The performance of the attention map at different placements in the middle flow of the XceptionNet architecture.}
        \vspace{2mm}
             \resizebox{\linewidth}{!}{%
    \begin{tabular}{|c|c|c|c|c|c|}
        \hline
        Map position & AUC & EER & TDR$_{0.01\%}$ & TDR$_{0.1\%}$ & PBCA \\ \hline
        Block1 & $99.82$ & $\textbf{1.69}$ & $71.46$ & $\textbf{92.80}$ & $83.30$\\ \hline
        Block2 & $\textbf{99.84}$ & $1.72$ & $67.95$ & $90.14$ & $87.41$\\ \hline
        Block3 & $99.50$ & $2.82$ & $49.06$ & $72.50$ & $88.14$\\ \hline
        Block4 & $99.64$ & $2.23$ & $\textbf{83.83}$ & $90.78$ & $88.44$ \\ \hline
        Block5 & $99.49$ & $2.62$ & $82.70$ & $89.03$ & $88.40$\\ \hline
        Block6 & $99.72$ & $2.28$ & $63.08$ & $86.02$ & $87.41$\\ \hline
        Block7 & $99.78$ & $1.79$ & $28.51$ & $88.98$ & $88.39$\\ \hline
        Block8 & $98.62$ & $4.42$ & $74.24$ & $79.95$ & $\textbf{88.96}$\\ \hline
    \end{tabular}
    }
    \label{tab:map_position}
    \vspace{-3mm}
\end{table}

{\small
\bibliographystyle{ieee_fullname}
\bibliography{egbib}
}

%\newpage
%\subfile{sec_7_supp.tex}

\end{document}